\documentclass[10pt,conference]{IEEEtran}

\usepackage[numbers, sort]{natbib}
\usepackage{comment}
\usepackage{balance}
\usepackage{amsmath}
\usepackage{hyperref}
\usepackage{cleveref}
\usepackage{mathtools}
\usepackage{multicol}
\usepackage{multirow}
\usepackage{xspace}
\usepackage{xcolor}
\usepackage{stfloats}
\usepackage{fancybox}
\usepackage{graphicx}
\usepackage{lipsum}
\usepackage{enumitem}
\usepackage{pbox}
\usepackage{colortbl}

\usepackage{svg}

\usepackage[labelfont={bf}, font={normalsize}]{caption}
\usepackage{subcaption}
\usepackage{fancyhdr, etoolbox}
\usepackage{fancyhdr}
\begin{document}

\fancypagestyle{firstpage}{
     \fancyhf{}
    \fancyhead[L]{}
    \fancyhead[C]{\fontsize{9}{14}\selectfont The 1st International Workshop on Interpretability and Robustness in Neural Software Engineering (InteNSE'23), \\ Co-located with ICSE'23, Melbourne, Australia}
    \fancyhead[R]{}
}

\makeatletter
\newcommand{\newlineauthors}{%
  \end{@IEEEauthorhalign}\hfill\mbox{}\par
  \mbox{}\hfill\begin{@IEEEauthorhalign}
}
\makeatother

\title{A Study of Variable-Role-based Feature Enrichment in Neural Models of Code}



\author{
    \IEEEauthorblockN{Aftab Hussain}
    \IEEEauthorblockA{
        \textit{ahussain27@uh.edu} \\
        University of Houston \\
        Houston, TX, USA
    }
    \and
    \IEEEauthorblockN{Md Rafiqul Islam Rabin}
    \IEEEauthorblockA{
        \textit{mrabin@uh.edu} \\
        University of Houston \\
        Houston, TX, USA
    }
    \and
    \IEEEauthorblockN{Bowen Xu}
    \IEEEauthorblockA{
        \textit{bowenxu@smu.edu.sg} \\
        Singapore Management University \\
        Singapore
    }
    
    \newlineauthors

    \IEEEauthorblockN{David Lo}
    \IEEEauthorblockA{
        \textit{davidlo@smu.edu.sg} \\
        Singapore Management University \\
        Singapore
    }

    \and
    
    \IEEEauthorblockN{Mohammad Amin Alipour}
    \IEEEauthorblockA{
        \textit{maalipou@central.uh.edu} \\
        University of Houston \\
        Houston, TX, USA
    }
}

\maketitle
\thispagestyle{firstpage}

\newcommand{\eg}{\textit{e.g.}\xspace}
\newcommand{\ie}{\textit{i.e.}\xspace}
\newcommand{\etal}{\emph{et al.}\xspace}
\newcommand{\resp}{\emph{resp.}\xspace}
\newcommand{\aka}{\emph{a.k.a.}\xspace}

\newcommand{\Part}[1]{\noindent\textbf{#1}}
\newcommand{\crossmark}{$\times$}
\newcommand{\Space}[1]{}

\newcommand{\FIX}[1]{\textbf{[\textcolor{red}{FIX: #1}]}}
\newcommand{\TODO}[1]{\textbf{[\textcolor{red}{TODO: #1}]}}

\newcommand{\approach}{\textsc{Skepticism}\xspace}
\newcommand{\DD}{Delta Debugging\xspace}
\newcommand{\ddmin}{ddmin\xspace}
\newcommand{\csnap}{\textsc{code2snapshot}\xspace}
\newcommand{\ttoken}{\textsc{Token}\xspace}

\newcommand{\CI}{Code Intelligence\xspace}
\newcommand{\Ci}{Code intelligence\xspace}
\newcommand{\ci}{code intelligence\xspace}

\newcommand{\mn}{\textsc{MethodNaming}\xspace}
\newcommand{\vm}{\textsc{VarMisuse}\xspace}
\newcommand{\vd}{\textsc{SoftVulDetection}\xspace}

\newcommand{\ctv}{Code2Vec\xspace}
\newcommand{\cts}{Code2Seq\xspace}
\newcommand{\rnn}{RNN\xspace}
\newcommand{\tra}{Transformer\xspace}
\newcommand{\cnn}{CNN\xspace}
\newcommand{\resNet}{ResNet\xspace}
\newcommand{\nmcs}{{neural models of code}\xspace}

\newcommand{\jlarge}{JavaLarge\xspace}
\newcommand{\pyg}{Py150G\xspace}
\newcommand{\clean}{\textsc{Clean}\xspace}
\newcommand{\norm}{\textsc{Normalized}\xspace}
\newcommand{\xalpha}{\textsc{Redacted}\xspace}

\newcommand{\mnp}{\textsc{MethodName}\xspace}

\newcommand{\JS}{\textsc{Java-Small}\xspace}
\newcommand{\JM}{\textsc{Java-Med}\xspace}
\newcommand{\JL}{\textsc{Java-Large}\xspace}
\newcommand{\JLR}{\textsc{Java-Large-Roles}\xspace}
\newcommand{\JLT}{\textsc{Java-Large-Transformed}\xspace}
\newcommand{\JLTR}{\textsc{Java-Large-Transformed-Roles}\xspace}
\newcommand{\JVS}{\textsc{Java-Vsmall}\xspace}
\newcommand{\JVSR}{\textsc{Java-Vsmall-Roles}\xspace}
\newcommand{\PY}{\textsc{Py150-Great}\xspace}
\newcommand{\SA}{\textsc{Sorting-Algorithm}\xspace}
\newcommand{\JTT}{\textsc{Java-Top10}\xspace}

\newcommand{\nmc}{{neural model of code}\xspace}
\newcommand{\ggnn}{\textsc{GGNN}\xspace}
\newcommand{\great}{\textsc{Great}\xspace}
\newcommand{\ctso}{\cts-$O$\xspace}
\newcommand{\ctsr}{\cts-$R$\xspace}
\newcommand{\ctsos}{\cts-$O_s$\xspace}
\newcommand{\ctsrs}{\cts-$R_s$\xspace}

\begin{abstract}
Although deep neural models substantially reduce the overhead of feature engineering, the features readily available in the inputs might significantly impact training cost and the performance of the models. In this paper, we explore the impact of an unsuperivsed feature enrichment approach based on variable roles on the performance of neural models of code. The notion of variable roles (as introduced in the works of Sajaniemi et al.~\cite{jorma,roles-web}) has been found to help students' abilities in programming. In this paper, we investigate if this notion would improve the performance of neural models of code. To the best of our knowledge, this is the first
work to investigate how Sajaniemi et al.'s concept of variable roles can affect neural models of code.  In particular, we enrich a source code dataset by adding the role of individual variables in the dataset programs, and thereby conduct a study on the impact of variable role enrichment in training the \cts model. In addition, we shed light on some challenges and opportunities in feature enrichment for neural code intelligence models. 
\end{abstract}


\section{Introduction}
\label{sec-intro}

In recent years, there has been significant growth in the research and development of AI models for software engineering tasks (referred to as \textit{code intelligence models} or \textit{\nmcs}) in large software companies like Facebook, Microsoft, and Google; for example, automatic code completion~\cite{aroma,lachaux20}, method name prediction~\cite{alon2018code2seq,alon2018code2vec}, bug prediction~\cite{hellendoorn2020global}, and even code generation from natural language (e.g. Github Copilot)~\cite{github2021copilot}. A key, costly, and often necessary factor that allows deep neural models in performing these tasks more accurately is scaling up: training massive models on huge datasets~\cite{dl-growing-cost} for extensive time periods -- putting significant stress on computation resources. For instance, it has been estimated that training Lachaux et al.’s~\cite{lachaux20} final published model with 32 GPUs costs around \$30,000, while training
the published GPT-3~\cite{gpt3} language model would cost more than \$2 million~\cite{dl-growing-cost}. These tough realities underscore the need for further investigation in approaches that can improve the performance of deep neural models, under more sustainable settings.

One approach that aims to help neural networks (NNs) learn faster and more effectively is \emph{feature enrichment}, which adds extra information in individual inputs, e.g., by explicating the data dependence relations between variables. 
On the other hand, such techniques are likely to benefit only up to the extent to which the added ``explicit'' knowledge brings in ``new'' knowledge to the dataset. For instance, if the added knowledge captures relations in the data that are already being learned by the neural model, the enrichment process is unlikely to bring substantial gains in the NN's performance. 

In this work, we evaluate the potential of an unsupervised source code feature enrichment approach for addressing the scalability bottleneck of deep neural models: enriching a given source code dataset with explicit variable role information, where a \textit{role} reflects the manner in which a variable interacts with other variables in a program. The notion of variable roles was introduced in the works of Sajaniemi et al.~\cite{jorma,roles-web}, who classified variables in programs into role categories based on how they are used (e.g., \emph{fixed value}, \emph{temporary variable}, \emph{stepper}, etc.). They applied the role concepts in programming learning tasks and found that roles facilitated the abilities of students to mentally process program information and apply programming steps in writing programs. Inspired by this work, we are interested in how data enrichment with variable role information can impact neural models of code. In particular, we seek to answer the following question: \emph{can explicitly injecting the information about variable roles to the input data help neural models of code achieve better performance and robustness with lesser effort spent on training?} 


To pursue this investigation, we (1) build a static analyzer to identify two common kinds of variable roles (\emph{steppers} and \emph{walkers}), and (2) augment the dataset with explicit roles information. We then train a popular code intelligence model, \cts~\cite{alon2018code2seq}, using the role-augmented dataset. Next, we compare our trained model against a pretrained version of \cts, released by the creators of the model, who had pretrained it with the original (unaugmented) dataset. In addition, we evaluate the models' robustness by testing them on noise-induced data (data in which variables in a program are randomly transformed). For both the role augmentation and noise induction steps, we use the semantic-preserving transformation technique~\cite{rabin2021generalizability}. Since we aimed to do an exploratory study, we chose \cts for its ease-of-use.

\noindent\textbf{Contributions}. This paper provides the following contributions:

\begin{itemize}[leftmargin=*]
\item We present a novel exploratory study to investigate the impact of augmenting variable roles information in a large code dataset on the performance and robustness of \cts.
\item We present and implement a technique to automatically detect certain roles of variables in programs, and consequently embed the roles information into variables in the input programs.
\item We provide insights on the usefulness of adding signals like roles to code datasets in using deep neural models to perform software engineering tasks.
\end{itemize}

\noindent\textbf{Paper Organization}. The rest of this paper is organized as follows: In Section~\ref{sec-method}, we discuss the methodology of our work, where we provide definitions of the variable roles we used, and present our variable role detection and augmentation approach. In Sections~\ref{sec-design} and~\ref{sec-res}, we present our experimental design and results, respectively. In Section~\ref{sec-disc}, we provide a discussion on the insights we gained from the results, giving some future directions. In Section~\ref{sec-threats}, elaborate on the threats to validity of our work. In Section~\ref{sec-rel}, we present some related literature. We conclude the paper in Section~\ref{sec-conc}.
\section{Methodology}
\label{sec-method}

In this section, we discuss our approach. We first, present the definitions of the two specific variable roles, which we studied in this work. Then we discuss how we added roles information to variables in a program in Subsection~\ref{subsec-aug-approach}. 

\subsection{Roles Definitions}
\label{subsec-roles}

Towards augmenting variable role information into programs, we focused on two common kinds of variable roles: \textit{steppers} and \textit{walkers}. We define these roles based on the work of Hermans~\cite{hermans2021}, who reintroduced Sajaniemi's cite variable role categories~\cite{roles-web,jorma}. In contrast to these previous works, we provide more refined definitions that are syntax-oriented and geared towards object-oriented programming languages like Java:

\begin{itemize}[leftmargin=*]
    \item \textit{Stepper variable}. A stepper variable (or \textit{stepper}) is a for-loop variable of numeric type, which iterates through a list of values via arithmetic-operation based updates, such as, increment operations (e.g., \texttt{i++}), or more complex operations (e.g., \texttt{size = size/2}). In the following piece of code, \texttt{i} is a stepper,
    
    \noindent \begin{center}
    \texttt{for (int \textcolor{blue}{i}=0 ; \textcolor{blue}{i}<5; \textcolor{blue}{i}++)\{...\}}
    \end{center}
    
   
    \item \textit{Walker variable}. A walker variable (or \textit{walker}) can be of two kinds: (1) an iterator object, that enables traversal through containers via APIs, e.g., \texttt{iter} in the following piece of code, 
           \noindent \begin{center}
    \texttt{while (\textcolor{blue}{iter}.hasNext())\{...\}}
    \end{center}
    or (2) an enhanced for-loop variable, e.g., \texttt{elem} in the following, 
    \noindent \begin{center}
    \texttt{for (String \textcolor{blue}{elem}: Elements) \{...\}}
    \end{center}
    
    Unlike steppers, walkers can only be used for sequential access and can be of any type. 
    
\end{itemize}

\subsection{Role Detection and Role Augmentation}
\label{subsec-aug-approach}

Figure~\ref{fig-role-proc} shows the overall workflow of the role detection and augmentation process. We first built a static analyzer, \textsc{Role Detector}, which is based on a popular Java parser library \texttt{javalang}\footnote{https://github.com/c2nes/javalang} written in Python.
Using \textsc{Role Detector} we detect steppers and walkers based on the definitions presented in Section~\ref{subsec-roles} from an input program. Then, for implementing role augmentation, we built \textsc{Role Augmenter}, which is based on \texttt{JavaMethodTransformer}\footnote{https://github.com/mdrafiqulrabin/tnpa-generalizability/}, a semantic-preserving program transformation tool written in Java~\cite{rabin2021generalizability}. \textsc{Role Augmenter} augments roles information obtained from \textsc{Role Detector} into variables in the input program; specifically, all instances of the variable name in the actual program are prefixed with the role type. Here is an example:

\begin{figure}[htbp]
  \centering
  \includegraphics[scale=0.55]{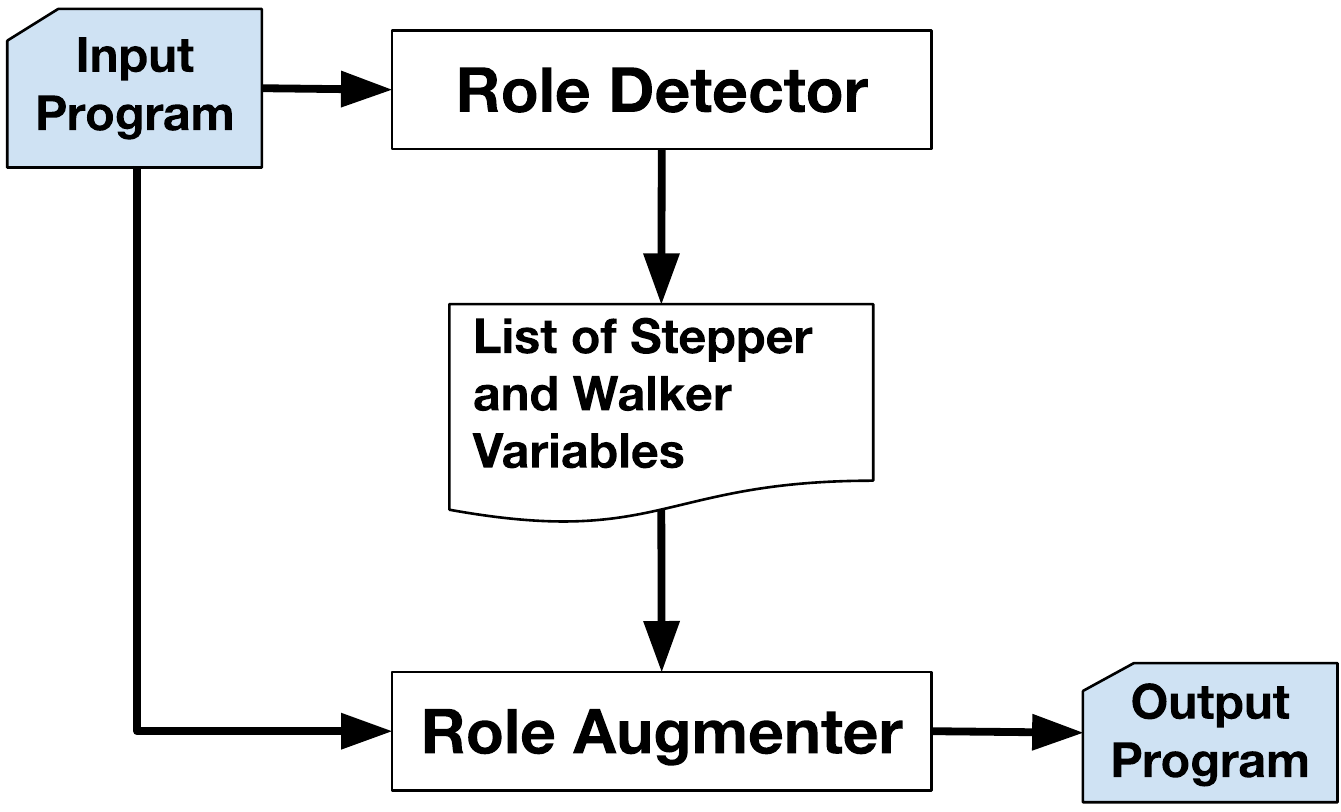}
  \caption{Role detection and augmentation workflow.}
  \label{fig-role-proc}
\end{figure}

\noindent \begin{center}
    \texttt{for (int count=0 ; count<10; count++)\{...\}}
\end{center}
\noindent \textsc{Role Detector} identifies \texttt{count} as a stepper in the above code. \textsc{Role Augmenter} then 
adds the prefix ``stepper\_'' to all instances of \texttt{count}, generating the following, 

\noindent \begin{flushleft}
    \texttt{for (int \textcolor{red}{stepper\_}count=0 ; \textcolor{red}{stepper\_}count<10; \textcolor{red}{stepper\_}count++)\{...\}}
\end{flushleft}



An alternative approach can be to implement both role detection and augmentation in a single pass using \texttt{JavaMethodTransformer}. However, our approach using~\texttt{javalang} for role detection is easier to implement as it provides an easy-to-read parse tree representation that can be checked for syntactic properties of steppers and walkers. In addition, our two-pass approach allows us to independently test the detection and augmentation phase more cleanly. 

Our variable roles augmentation gives potential hints to the model about both the structure and semantics of the code, where specific tokens are added for variables in certain kinds of program structures (in our case, loops). Code models can be very dependent on variable names (e.g., CodeBERT)~\cite{multi-lingual-training}, while others may have more reliance on the structure of the program~\cite{bowen-icse23}. \cts encodes paths in the abstract syntax tree, in addition to encoding tokens into subtokens, and has been found to perform well even for unseen inputs. Our investigation thus throws light on whether \cts can leverage any benefit, if there is useful information in the input tokens. 

\section{Experimental Design}
\label{sec-design}

In this section, we discuss the datasets, models, and the evaluation approach we used in our experiments.

\subsection{Datasets}

We deployed our role detection and augmentation techniques on the \JL methods dataset~\cite{alon2018code2seq} to generate the role-augmented dataset, \JLR. Each sample in these datasets corresponds to a Java method. Table~\ref{tab-jl-jlr} shows the number of samples in each split of the datasets, the number of augmented samples in each split of \JLR  (shown in parentheses), and the number of variables augmented as steppers and walkers. (No steppers and walkers were detected in the un-augmented samples.) 

{\renewcommand{\arraystretch}{1.3}
\begin{table}[]
\caption{Size of \JL (JL) and \JLR (JLR) datasets, and number of variables augmented in these datasets. In parentheses, we provide the number of \JL methods augmented with roles information to create the \JLR dataset.}
\centering
\setlength{\tabcolsep}{2pt}
\scalebox{1.05}{\begin{tabular}{l|ccc|ccc}
\hline
 \multirow{2}{*}{\textbf{Dataset}}       & \multicolumn{3}{c|}{\underline{\textbf{No. of methods in each split}}}                                        & \multicolumn{3}{c}{\underline{\textbf{No. of vars. augmented}}}  \\
 & Test & Train & Val & \textit{Steppers}  & \textit{Walkers} & Total \\  \cline{1-7}  \hline  \hline
JL      & 370,930                  &      13,376,784                     &        274,604                         &                         N/A      &        N/A           &      N/A   \\ \arrayrulecolor{gray}\hline \arrayrulecolor{black}
 \multirow{2}{*}{JLR}    & 370,930         &       13,376,784                  &            274,604                     &          \multirow{2}{*}{624,701}                     &        \multirow{2}{*}{813,334 }            &     \multirow{2}{*}{1,438,035}   \\
    & (32,358)         &        (1,063,868)                    &            (20,933)                     &                        &                 &        \\\hline
\end{tabular}}
\label{tab-jl-jlr}
\end{table}
}

\subsection{Models}

We study the \cts model trained from scratch using the \JLR dataset. We compare our trained version with the pretrained version released by the model's creators~\cite{alon2018code2seq}. We refer to these two versions as \ctsr (\cts-Roles) and \ctso (\cts-Original). In total, 50 epochs were spent to generate \ctsr (i.e., the best performing model found in terms of $F$-1 score after 50 epochs of training). The pretrained released version, \ctso, was obtained after 52 epochs of training. 


\subsection{Prediction Task}

The prediction task is to predict the method name of a function body, when only provided the function body. This task is widely considered in many prior works, e.g.~\cite{alon2018code2vec,learning-sem-embedd, infercode}. 

\subsection{Model Architecture}
\label{subsec-subj-model}

\cts~\cite{alon2018code2seq} uses an encoder-decoder architecture. The encoder encodes paths in the abstract syntax tree (AST) of programs, where each path corresponds to a sequence of nodes in the AST. \cts treats the input function code as a sequence of tokens, which are further split into sub-tokens. The decoder uses attention mechanism~\cite{vaswani2017attention} to extract features from relevant paths and predicts sub-tokens of a target sequence in order to generate the output, which in our study is the method name. 

\subsection{Evaluation Test Sets}

We evaluated the models using the eight test sets obtained in the way shown in Figure~\ref{fig:evaluation}. 

The ``Original'' and ``Original,
Roles-added'' sets directly correspond to the test sets in \JL and \JLR, respectively. The ``Original,
Roles-added'' test set consists of both augmented and unaugmented methods. The ``Original'' set consists of all methods in the ``Original, Roles-added'' set, in un-augmented form. These test sets help us evaluate whether training a model with role augmented data affects its predictions for diverse inputs, with and without the variable roles we considered. 

To more accurately gauge the effects of role augmentation on the models, we also tested the models with filtered versions of each of the above test sets, where we only kept augmented methods. Thus, ``Original, Roles-added, Filtered'' set only consists of augmented methods, and ``Original, Filtered'' set consists of all methods in ``Roles-added, Filtered'' set, in un-augmented form. 

Finally, we evaluated the models on the face of semantically-transformed test data to examine their robustness; we applied variable name transformations randomly on the \JL and \JLR test sets using the approach in~\cite{rabin2021generalizability} to obtain transformed test sets. In the transformation, all occurrences of a randomly picked variable in a program was changed to a generic variable name, ``varN'', where ``N'' is an integer. This transformation thus strips away any semantic meaning in a variable, and thus is a form of noise induction. 

\noindent We summarize all the test sets in Table~\ref{tab-jlt-jltr}.

{\renewcommand{\arraystretch}{1.3}
\begin{table}[]
\caption{Sizes of both unfiltered and filtered \JL (JL), \JLR (JLR), \JLT (JLT), \JLTR (JLTR) test sets. (F) corresponds to the filtered versions.}
\centering
\setlength{\tabcolsep}{2pt}
\scalebox{1.15}{\begin{tabular}{c|c|c}
\hline
Type & \textbf{Test Set}  & \textbf{No. of methods}     \\ 
                                                    \hline
                                                  \hline
\multirow{4}{*}{Untransformed} &  JL Test       &      370,930                          \\
&JLR Test       &      370,930                    \\
&JL Test (F)      &      32,358                        \\
&JLR Test (F)     &      32,358                    \\
\arrayrulecolor{gray}\hline \arrayrulecolor{black}
\multirow{4}{*}{Transformed} & JLT Test       &      916,611                       \\
&JLTR Test     &       916,611       \\ 
&JLT Test (F)      &      261,832                         \\
&JLTR Test (F)    &       261,832     \\ 
\hline
\end{tabular}}
\label{tab-jlt-jltr}
\end{table}
}

\begin{figure}[htbp]
  \centering
  \includegraphics[scale=0.55]{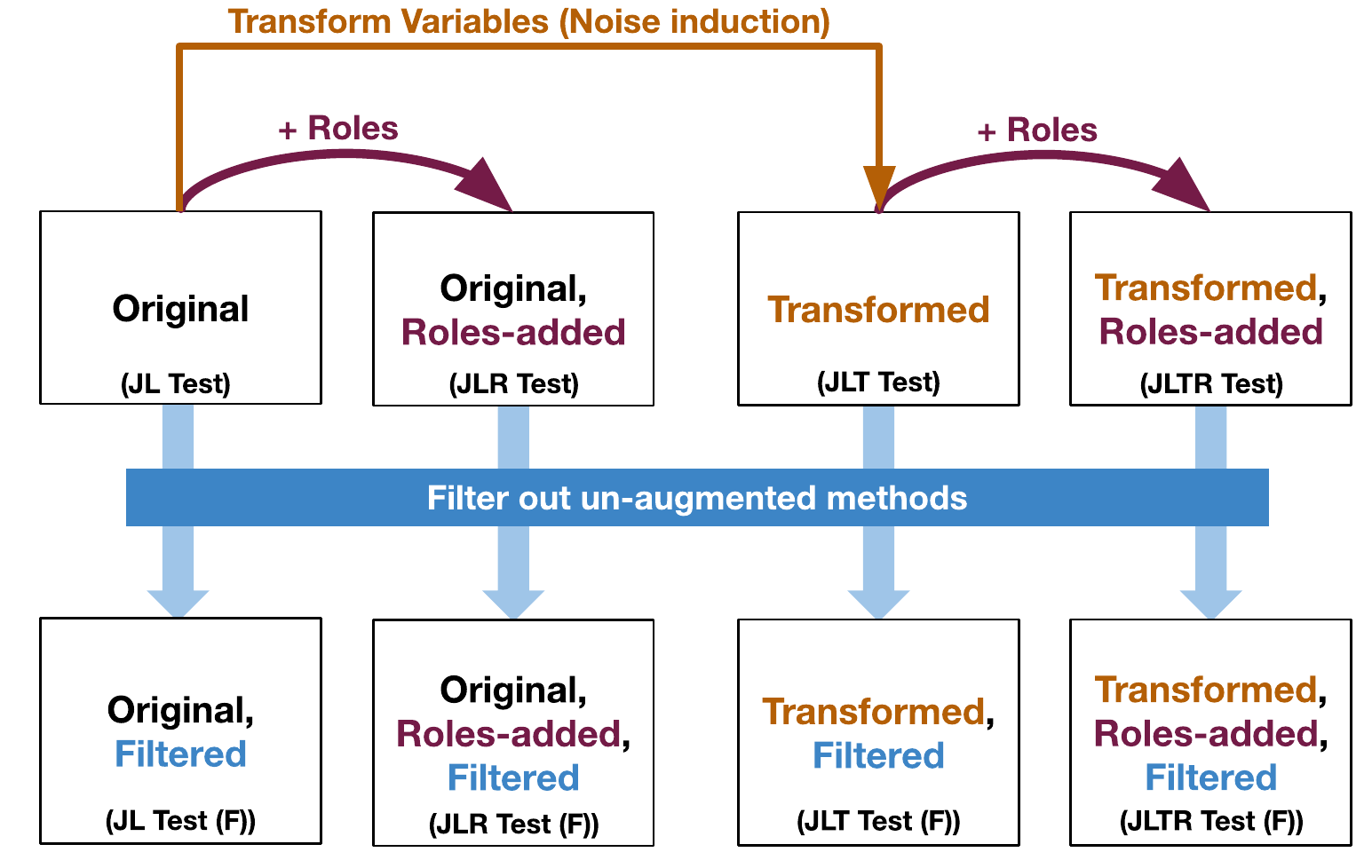}
  \caption{The approach for obtaining the eight test sets that were used for evaluating the models. The ``Original'' and ``Original, Roles-added'' test sets correspond to the test sets in  \JL and \JLR (JL Test and JLR Test), respectively.}
    \label{fig:evaluation}
\end{figure}

\subsection{Metrics} 

In order to evaluate the model's predictions, we use the same metrics of precision, recall, and F1-score used at the sub-token level, as in~\cite{rabin2021generalizability}. These metrics are typically used for method name prediction tasks~\cite{alon2018code2seq,alon2018code2vec}. 

\section{Results}
\label{sec-res}

In this section, we present the results of our experiments, where we seek to answer the following research questions:

\begin{itemize}
    \item[RQ.1] How does role augmentation impact \cts's \textbf{effectiveness} in making predictions? (Subsection~\ref{subsec-rq1})
    \item[RQ.2] How does role augmentation impact \cts's \textbf{robustness} in making predictions? (Subsection~\ref{subsec-rq2})
\end{itemize}

\subsection{RQ.1: Impact of Role Augmentation on the Effectiveness of \cts}
\label{subsec-rq1}

To evaluate the overall effectiveness of \ctsr and \ctso, we tested the models with the untransformed test sets, the results of which are shown in Table~\ref{tab-perf} (Test type (a)). We observed that \ctsr attained slightly better precision values with all four test sets. On the contrary, the recall values were slightly better for \ctso compared to \ctsr for three of the untransformed test sets. Overall, however, the $F_1$-scores were very similar between the two models, for all the test sets. 

To gain further insight into the impact of role-based feature enrichment, we wanted to observe the behaviour of \cts when it is trained with a dataset containing a greater share of role-augmented samples. 
We thus curated the dataset \JVSR by randomly extracting a subset of only role-augmented samples in \JLR (In total, 157,046 train set examples and 19,931 validation set examples). Thus 100\% of the samples in \JVSR are role-augmented. In addition, we constructed an equivalent un-augmented version, \JVS, which consists of all samples in \JVSR in un-augmented form. 

The smaller versions of the datasets made it feasible to train with both the augmented and un-augmented datasets for longer epochs. We thus separately trained \cts from scratch on \JVS and \JVSR for 65 epochs, and saved the best models in each train session based on $F_1$-score; the saved models are referred to as \ctsos and \ctsrs, respectively. 

We evaluated the overall effectiveness of \ctsos and \ctsrs using the same test sets used for evaluating \ctso and \ctsr. The results are shown in Table~\ref{tab-perf-vs} (Test type (a)). We observed very similar marginal differences for \ctsos and \ctsrs, as we did for \ctso and \ctsr. In precision, \ctsrs was slightly better with all four test sets. However, overall, the $F_1$-scores were very similar between the two models, for all the test sets. 

\noindent \textbf{Observation.} \textit{Overall, the experiment results demonstrate that the impact of role augmentation is \textbf{indistinguishable} with respect to \cts's performance in making method name predictions. In other words, role augmentation neither boosted nor harmed the performance significantly.}

{\renewcommand{\arraystretch}{1.2}
\begin{table}[]
\caption{Precision, recall, and $F_1$ scores for predictions by the two trained \cts models on the eight test sets obtained from \JL and \JLR as shown in Figure~\ref{fig:evaluation}. $O$ and $R$ indicate the values for \cts-$O$ and \cts-$R$, respectively. The numbers in type (a) and (b) correspond to the results for untransformed and transformed test sets respectively.}
\centering
\setlength{\tabcolsep}{2pt}
\scalebox{1.05}{
\begin{tabular}{c|l|cc|cc|cc}
\hline
\multirow{2}{*}{\textbf{Type}}  &\multirow{2}{*}{\textbf{Test Set}}  & \multicolumn{2}{c|}{\textbf{Precision}}          & \multicolumn{2}{c|}{\textbf{Recall}}                                    & \multicolumn{2}{c}{\textbf{$F_1$-Score}}                                   \\
             &     & $O$   & $R$  & $O$   & $R$   & {$O$}   & $R$ \\ \hline \hline
\multirow{4}{*}{(a)}  & JL     & \multicolumn{1}{c}{0.636} & \multicolumn{1}{c|}{0.643}
                            & \multicolumn{1}{c}{0.541} & \multicolumn{1}{c|}{0.536} 
                            & \multicolumn{1}{c}{0.585} & \multicolumn{1}{c}{0.584}        \\
  & JL  (F)   & \multicolumn{1}{c}{0.458} & \multicolumn{1}{c|}{0.462} 
                            & \multicolumn{1}{c}{0.375} & \multicolumn{1}{c|}{0.365}
                            & \multicolumn{1}{c}{0.412} & \multicolumn{1}{c}{0.408}     \\
  & JLR            & \multicolumn{1}{c}{0.636} & \multicolumn{1}{c|}{0.643} 
                        & \multicolumn{1}{c}{0.54} & \multicolumn{1}{c|}{0.536} 
                        & \multicolumn{1}{c}{0.584} & \multicolumn{1}{c}{0.584}      \\
  & JLR (F)   & \multicolumn{1}{c}{0.447} & \multicolumn{1}{c|}{0.462}
                        & \multicolumn{1}{c}{0.36} & \multicolumn{1}{c|}{0.367} 
                     & \multicolumn{1}{c}{0.399} & \multicolumn{1}{c}{0.409}     \\  \arrayrulecolor{gray}\hline \arrayrulecolor{black}
\multirow{4}{*}{(b)}  & JL-Transformed              & \multicolumn{1}{c}{0.493} & \multicolumn{1}{c|}{0.497} 
                         & \multicolumn{1}{c}{0.401} & \multicolumn{1}{c|}{0.398} 
                        & \multicolumn{1}{c}{0.442} & \multicolumn{1}{c}{0.442}      \\
&JL-Transformed  (F)    & \multicolumn{1}{c}{0.418} & \multicolumn{1}{c|}{0.42} 
                        & \multicolumn{1}{c}{0.346} & \multicolumn{1}{c|}{0.34} 
                        & \multicolumn{1}{c}{0.379} & \multicolumn{1}{c}{0.376}        \\
&JLR-Transformed             & \multicolumn{1}{c}{0.493} & \multicolumn{1}{c|}{0.497}
                        & \multicolumn{1}{c}{0.398} & \multicolumn{1}{c|}{0.398} 
                        & \multicolumn{1}{c}{0.44} & \multicolumn{1}{c}{0.442}  \\
&JLR-Transformed  (F)   & \multicolumn{1}{c}{0.415} & \multicolumn{1}{c|}{0.42} 
                            & \multicolumn{1}{c}{0.336} & \multicolumn{1}{c|}{0.341} 
                            & \multicolumn{1}{c}{0.372} & \multicolumn{1}{c}{0.377} \\
 
                     \hline
\end{tabular}
}
\label{tab-perf}
\end{table}}

{\renewcommand{\arraystretch}{1.2}
\begin{table}[]
\caption{Precision, recall, and $F_1$ scores for predictions by the two trained \cts models on the eight test sets obtained from \JVS and \JVSR as shown in Figure~\ref{fig:evaluation}. $O_S$ and $R_S$ indicate the values for \cts-$O_S$ and \cts-$R_S$, respectively. The numbers in type (a) and (b) correspond to the results for untransformed and transformed test sets respectively.}
\centering
\setlength{\tabcolsep}{2pt}
\scalebox{1.05}{
\begin{tabular}{c|l|cc|cc|cc}
\hline
\multirow{2}{*}{\textbf{Type}}  &\multirow{2}{*}{\textbf{Test Set}}  & \multicolumn{2}{c|}{\textbf{Precision}}          & \multicolumn{2}{c|}{\textbf{Recall}}                                    & \multicolumn{2}{c}{\textbf{$F_1$-Score}}                                   \\
             &     & $O_S$   & $R_S$  & $O_S$   & $R_S$   & $O_S$   & $R_S$ \\ \hline \hline
\multirow{4}{*}{(a)}  & JL             & \multicolumn{1}{c}{0.465} & \multicolumn{1}{c|}{0.467}
                                    & \multicolumn{1}{c}{0.307} & \multicolumn{1}{c|}{0.306}
                                & \multicolumn{1}{c}{0.37} & \multicolumn{1}{c}{0.37} \\
  & JL  (F)                  & \multicolumn{1}{c}{0.356} & \multicolumn{1}{c|}{0.358}
                            & \multicolumn{1}{c}{0.256} & \multicolumn{1}{c|}{0.255}
                            & \multicolumn{1}{c}{0.298} & \multicolumn{1}{c}{0.298}   \\
  & JLR                      & \multicolumn{1}{c}{0.465} & \multicolumn{1}{c|}{0.466}
                            & \multicolumn{1}{c}{0.306} & \multicolumn{1}{c|}{0.306}
                            & \multicolumn{1}{c}{0.369} & \multicolumn{1}{c}{0.37}\\
  & JLR (F)                 & \multicolumn{1}{c}{0.355} & \multicolumn{1}{c|}{0.356}
                            & \multicolumn{1}{c}{0.256} & \multicolumn{1}{c|}{0.258}
                            & \multicolumn{1}{c}{0.297} & \multicolumn{1}{c}{0.299}
     \\  \arrayrulecolor{gray}\hline \arrayrulecolor{black}
\multirow{4}{*}{(b)}  & JL-Transformed              & \multicolumn{1}{c}{0.362} & \multicolumn{1}{c|}{0.361}
                                            & \multicolumn{1}{c}{0.239} & \multicolumn{1}{c|}{0.241}
                                        & \multicolumn{1}{c}{0.288} & \multicolumn{1}{c}{0.289}      \\
&JL-Transformed  (F)    & \multicolumn{1}{c}{0.337} & \multicolumn{1}{c|}{0.335}
                        & \multicolumn{1}{c}{0.24} & \multicolumn{1}{c|}{0.236}
                        & \multicolumn{1}{c}{0.281} & \multicolumn{1}{c}{0.277}       \\
&JLR-Transformed            & \multicolumn{1}{c}{0.362} & \multicolumn{1}{c|}{0.36}
                            & \multicolumn{1}{c}{0.239} & \multicolumn{1}{c|}{0.241}
                            & \multicolumn{1}{c}{0.288} & \multicolumn{1}{c}{0.289}  \\
&JLR-Transformed  (F)  & \multicolumn{1}{c}{0.336} & \multicolumn{1}{c|}{0.333}
& \multicolumn{1}{c}{0.239} & \multicolumn{1}{c|}{0.238}
& \multicolumn{1}{c}{0.28} & \multicolumn{1}{c}{0.278} \\
                     \hline
\end{tabular}
}
\label{tab-perf-vs}
\end{table}}

\subsection{RQ.2: Impact of Role Augmentation on the Robustness of \cts}
\label{subsec-rq2}

To investigate the impact of role augmentation on the robustness of \cts, we tested \ctso and \ctsr on transformed data, the results of which are summarized in Table~\ref{tab-perf} (Test type (b)).  
We see the $F_1$-scores values of both \ctso and \ctsr dropped for the transformed test sets, with the difference between them being marginal, just as was seen with the untransformed tests. Overall, \ctso and \ctsr exhibited almost the same degrees of robustness (produced almost the same performance) with the transformed test sets.

Similar to the approach for evaluating the effectiveness of \cts, we also tested the \ctsos and \ctsrs models (refer Subsection~\ref{subsec-rq1}), this time with the transformed test sets. A similar trend of marginal differences between the scores for \ctsos and \ctsrs were also observed for these noise-induced test sets.  

\noindent \textbf{Observation.} \textit{Overall, the experimental findings show that the impact of role augmentation is \textbf{indistinguishable} with respect to \cts's robustness in making method name predictions. In other words, role augmentation neither boosted nor harmed the robustness of \cts significantly.}

\section{Discussion}
\label{sec-disc}








The results of our experiments are negative. Our feature-enrichment scheme did not significantly impact \cts. One reason for this observation could be that since variable roles are determined from the way variables are syntactically used, \cts may already be capable of capturing the surrounding structural context of a certain variable role, and this may explain the largely similar performances seen by \ctso and \ctsr. However, more investigations may be necessary to determine how much benefit does explicitly inserting the role in a variable name add to the predictions of the models, in cases where the role is less apparent from just the structure of the code (e.g., \textit{fixed-value} variables, i.e., constants, may be used in significantly different ways in code). 

\noindent \textit{Thus, towards investigating the impact of variable roles in neural program models that capture code structure, there is a need to distinguish between variables with roles that are more strictly reliant on code structure, e.g., steppers, and those that can be more flexibly used, despite playing the same role, e.g., fixed-value (constant) variables.}



Another direction that needs further investigation is towards understanding how capable are the models in exploiting enriched features, as it is ``possible'' that the models may have an intrinsic incapability in exploiting enriched features in datasets.
Separate investigations can be carried out in this direction for models that are more reliant on code semantics (e.g., large language-model-based deep neural networks like CodeBERT~\cite{codebert} and CodeGPT~\cite{codegpt}).

Finally, one more reason for the similarity in performances of \ctso and \ctsr could be the fact that only two variable roles were augmented, an aspect of our experiment we discuss in more detail in the next section.

\section{Threats to Validity}
\label{sec-threats}

Although we considered two common types of variable roles as our first attempt, still, they only covered 8\% of the samples of the whole dataset  (1,117,159 of the approximately 14 million methods). In the future, we plan to extend this work by considering more variable roles to further mitigate this threat.



Furthermore, the stepper variable names are not as diverse. Most steppers were seen to be the variables \texttt{i} and \texttt{j}: of 624,701 steppers detected in the entire \JL dataset, 475,383 (76.1\%) were \texttt{i}, and 47,221 (7.56\%) were \texttt{j}. Hence, adding the ``stepper'' keyword may not have had a significant effect in terms of adding additional information towards learning in the experiments. For future explorations, more of Sajaniemi's roles~\cite{jorma} could be added, which may require more complex static analysis techniques (e.g., data-flow analysis for detecting \textit{most-wanted-holder} and \textit{gatherer} variables~\cite{roles-web}.).

\section{Related Works}
\label{sec-rel}

In this section, we discuss two branches of works that have also focused on improving the performance of \nmcs: 

\textbf{Code Modeling.} There are numerous works that have tried to use different code representations in developing better performing neural program models.
Some early works used natural language processing models to capture textual patterns in code, without capturing structural information~\cite{gupta2017deepfix, pu2016sk_p}. Better performances were achieved by approaches that leveraged tree- and graph-forms of source code to better grasp code structure~\cite{alon2018code2seq,alon2018code2vec,maddison2014structured,mou2016convolutional,graphcodebert}. E.g., GraphCodeBERT has shown that representing variables with a static data flow graph and augmenting the graph with original source code can improve the transformer-based model's comprehension of code.~\cite{graphcodebert}.

In~\cite{chirkova2021embeddings}, the author developed dynamic embeddings, a recurrent mechanism that adjusts the learned semantics of the variable when it obtains more information about the variable's role in the program. They show that using the proposed dynamic embeddings significantly improves the performance of the recurrent neural network, in code completion and bug fixing tasks. In contrast to these techniques, our approach directly adds variable semantic information, as derived from Sajaniemi's pedagogical notion of variable roles~\cite{jorma,roles-web}, into the raw training source code dataset without adding a new representation of the source code (we only renamed the variables).
 
 \textbf{Feature Enrichment.} 
Allamanis et al.~\cite{allamanis2015suggesting} showed that adding features that capture global context can increase the performance of a model. Rabin et al.~\cite{rabin2020demystifying} found that code complexity features can improve the classification performance of some labels up to about 7\%. While this work focused on extracting a set of handcrafted features for better transparency, we study how feature enrichment affects in model's training behavior. Recent studies have shown that state-of-the-art models heavily rely on variables~\cite{compton2020embedding, multi-lingual-training}, specific tokens~\cite{rabin2021sivand}, and even structures~\cite{rabin2021code2snapshot}. Chen et al.~\cite{varclr} focus on semantic representations of program variables, and study how well models can learn similarity between variables that have similar meaning (e.g., \texttt{minimum} and \texttt{minimal}). Ding et al.~\cite{disco} explore the problem of learning functional similarities (and dissimilarities) between codes, towards which they rename variables to inject variable-misuse bugs in order to generate buggy programs that are structurally similar to benign ones. Neither of these works investigated or deployed variable-role based augmentation, as was done in this work. 







\section{Conclusion}
\label{sec-conc}

In this paper, we investigated the impact of explicitly adding variable role information in code datasets on the performance of \cts. To the best of our knowledge, this is the first work to evaluate the impact of Sajaniemi et al.'s notion of variable roles, a concept that was found to help students learn programming, to neural models of code. The work presents guidelines and challenges on enriching source code datasets for using code intelligence models more productively, encouraging the development of a systematic framework to investigate how to provide such models meaningful information to enable them to learn faster and perform better.

\section*{Acknowledgements}

We would like to thank our paper's reviewers from the InteNSE 2023 Program Committee for their valuable feedback towards the revision of this paper.

\balance
\bibliography{sample-base,ref}
\bibliographystyle{IEEEtranN}
\end{document}